# AUTOMATIC MOBILITY ANALYSIS OF PARALLEL MECHANISMS: AN ALGORITHM APPROACH BASED ON POSITION AND ORIENTATION CHARACTERISTIC EQUATIONS

**Xiaorong Zhu**
School of Mechanical Engineering, Changzhou University, Changzhou 213016, China
zxr@cczu.edu.cn

**Huiping Shen***
School of Mechanical Engineering, Changzhou University, Changzhou 213016, China
shp65@126.com

**Chengqi Wu**
School of Mechanical Engineering, Changzhou University, Changzhou 213016, China
753272704@qq.com

**Damien Chablat**
CNRS, Laboratoire des Sciences du Numérique de Nantes, UMR 6004 Nantes, France
Damien.Chablat@cnrs.fr

**Tingli Yang**
School of Mechanical Engineering, Changzhou University, Changzhou 213016, China
yangtl@126.com

## ABSTRACT

The determination of the mobility of parallel mechanisms (PM) is a fundamental problem. An automatic and intelligent analysis platform will be a significant tool for the design and optimization of mechanical systems. Based on the theory of position and orientation characteristics (POC) equations, a systematic approach to computer-aided mobility analysis of PMs is presented in this paper. First, a digital model for topological structures which has a mapping relationship with position and orientation characteristics of mechanism is proposed. It describes not only the dimension of the motion output, but also gives the mapping relationship between the output characteristic and the axis of the kinematic joints. Secondly, algorithmic rules are established that convert the union and intersection operations of POC into the binary logical operations and the automatic analysis of POC are realized. Then, the algorithm of the automatic mobility analysis of PMs and its implementation with VC++ are written .The mobility and its properties (POC) will also be analyzed and displayed automatically after introducing by users of the data of topological structures representation. Finally, typical examples are provided to show the effectiveness of the software platform.

*Corresponding author

## INTRODUCTION

Mobility analysis is the basis for the research on mechanical synthesis, kinematics and dynamics analysis. However, it is often difficult due to the fact of dealing with linear dependency for complex parallel mechanisms (PMs). With the development of computer technology, it would be very helpful to develop software for the mobility analysis. To evaluate the PMs mobility, several methods have been developed based on the screw theory [1, 2], the Lie group [3~5], the Position and Orientation Characteristics (POC) theory [6, 7] and the geometric algebra [8].

The approach upon screw theory has been successfully applied in many PMs. The calculation mainly involves the linear solution of screws which has the potential advantage of automatic analysis. However, it is difficult to obtain screws automatically. By establishing a coordinate system for a reference leg, Cao Wen'ao [9] realized an automatic analysis of the mobility of PMs.

The mobility analysis of the Lie group is based on the multiplication and intersection of displacement subgroup/submanifold. There are too many rules involved (over 107 rules) [4], and not suitable for programming. The other related efforts can also be found in recent works [10, 11]. However, most of the existing methods rely on manual and are inefficient. So, the study of automatic mobility analysis can provide an effective and practical means for designers.

The mobility analysis method based on the POC theory has a clear formula and judgment criteria that are easy to use and program [6]. To solve these problems, based on POC theory, a general computer-assisted platform for the analysis of PMs mobility will be written in this document. A digital model is proposed for topological structure that allows the POC to be mapped. An algorithm is established for calculation of POC of legs and PMs. Then, the principle and algorithm of PM mobility analysis are studied and software is developed. Finally, typical examples of parallel robots are used to show the efficiency of the software.

## THEORY OF POC-BASED

### Definition and related equations of POC

*Definition of POC*

To describe the relative motion characteristics between any two components in a mechanism, the POC is defined in [6] as

$$M = \begin{pmatrix} t^{\xi_{t1}} \text{ (dir.)} \\ r^{\xi_{r1}} \text{ (dir.)} \end{pmatrix} \quad (1)$$

$$\xi = \xi_t + \xi_r \leq F \quad (2)$$

where M is the POC, $\xi$ is rank of the POC (i.e. number of independent elements), dir means the direction of output axis, F is the DOF of the mechanism.

*POC equation for serial mechanisms*

The POC of a serial mechanism is

$$M_{Li} = \bigcup_{i=1}^{m} M_{Ji} \quad (3)$$

where $M_{Ji}$ is the POC of the $i^{th}$ kinematic joint, m is the number of kinematic joints, ∪ is "Union" operation [6].

*POC equation for PMs*

The POC of a PM is

$$M_{pa} = \bigcap_{j=1}^{\nu+1} M_{Lj} \quad (4)$$

where $M_{pa}$ is the POC of the moving platform, $\nu$ is the number of independent loops, $M_{Lj}$ is the POC of the $j^{th}$ leg for the same base point O', ∩ is "intersection" operation [6].

### Equation of mobility

A PM with ($\nu$+1) legs can be considered as a combination of $\nu$ independent loops (SLC). The structure composition of PMs based on SLC [6] is shown as Fig.1. Two legs are chosen to form the first independent loop (SLC$_1$), and the mobile platform is considered an active component of the loop; Then, SLC$_1$ is regard as a whole (an equivalent sub-PM), and combined with another leg to form the second independent loop (SLC$_2$); and in the same way, ($\nu$+1)$^{th}$ leg combined with SLC$_{(\nu-1)}$ to form the $\nu^{th}$ independent loop (SLC$_\nu$).

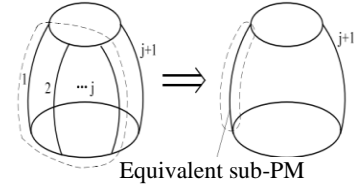

Equivalent sub-PM

FIGURE 1. THE BASIC INDEPENDENT LOOPS COMPOSED OF PMS

From this point, the number of mobility, i.e., the degree of freedom(DOF) of a PM can be calculated by Eq.(5), and the POC of the moving platform can be used to convey the property of mobility.

$$F = \sum_{i=1}^{m} J_i - \sum_{j=1}^{\nu} \xi_{L_j} = \sum_{j=1}^{\nu+1} f_j - \sum_{j=1}^{\nu} \xi_{L_j} \quad (5)$$

where,

$$\xi_{L_j} = \dim.\left(\left(\bigcap_{i=1}^{j} M_{b_i}\right) \cup M_{b_{(j+1)}}\right) \quad (6)$$

$$\nu = m - n + 1 \quad (7)$$

where F is the DOF of a PM, $J_i$ is the DOF of the $i^{th}$ joint, $f_j$ is the DOF of all joints in the $j^{th}$ leg, n is the number of components, $\xi_{Lj}$ is the number of independent displacement equations of the $j^{th}$ loop, $\bigcap_{i=1}^{j} M_{b_i}$ is the POC of the equivalent sub-PM composed by the front j legs, $M_{L(j+1)}$ is the POC of the (j+1)$^{th}$ leg.

### Key technologies for automatic mobility analysis

① The topological description of PMs is one of the most significant issues. It should include complete topological information of PMs with the most concise representation so that it can be recognized and extracted automatically.

② POC description for a leg and for a PM is a fundamental problem in automatic mobility analysis. Although the symbolic description as in Eq.(1) is good for expressing the geometric meaning, it is difficult to realize the transitivity of joint orientation relation.

③ The most critical issue is how to establish an algorithm for POC. In essential, the calculation of POC is to deal with linear dependency in complex PMs.

## REPRESENTATION OF MECHANISM AND ITS POC

### Description of the legs and their POC

Only simple legs constituted by single-DOF joints R and P are considered here. In general, joints in a PM are denoted by special symbols: R for revolute joint and P for prismatic joint.

In a leg, geometric relationship between joints is divided into six types: parallel, vertical, coaxial, spatial common point, coplanar and arbitrary, denoted by "⊥", "∥", "/", "*", "#" and "-" respectively.

*Digital matrix of leg*

A kind of ordered topology matrix (L) is developed for representing the topological structure of a leg in a PM. The diagonal element $J_i$ represents the type of pais from the fixed platform to the moving platform. Non-diagonal element $N_{ij}$ denotes the geometric relationship between joint i and joint j. The ordered topology matrix (L) is

$$L = \begin{bmatrix} J_1 & L & N_{1i} & L & N_{1f} \\ M & O & & & M \\ N_{i1} & & J_i & & N_{if} \\ M & & & O & M \\ N_{f1} & L & N_{fi} & L & J_f \end{bmatrix} \quad (8)$$

For convenience of programming, $N_{ij}$ (6 kinds of geometric relationship above-mentioned) is transformed into number 1~5, or 0 respectively. $J_i$ (R, P) is represented by 8 or 9. Thus, a leg can be expressed as a decimal matrix. Table 1 gives four typical legs and their matrices.

TABLE 1. TOPOLOGY MATRIX OF LEGS

| No. | Schematic diagram | Topology Matrix | POC matrix |
|---|---|---|---|
| (a) UP leg (RRP) | | $L = \begin{bmatrix} 8 & 2 & 2 \\ 2 & 8 & 2 \\ 2 & 2 & 9 \end{bmatrix}$ | $M_{La} = \begin{bmatrix} 0 & 0 & 1 \\ 1 & 1 & 0 \end{bmatrix}$ |
| (b) RRC leg (RRRP) | | $L = \begin{bmatrix} 8 & 1 & 1 & 1 \\ 1 & 8 & 1 & 1 \\ 1 & 1 & 8 & 1 \\ 1 & 1 & 1 & 9 \end{bmatrix}$ | $M_{Lb} = \begin{bmatrix} 3 & 0 & 0 & 0 \\ 1 & 0 & 0 & 0 \end{bmatrix}$ |
| (c) PRRRR leg | | $L = \begin{bmatrix} 9 & 2 & 2 & 1 & 1 \\ 2 & 8 & 1 & 2 & 2 \\ 2 & 1 & 8 & 2 & 2 \\ 1 & 2 & 2 & 8 & 1 \\ 1 & 2 & 2 & 1 & 8 \end{bmatrix}$ | $M_{Lc} = \begin{bmatrix} 3 & 0 & 0 & 0 & 0 \\ 0 & 1 & 0 & 1 & 0 \end{bmatrix}$ |
| (d) UPS leg (RRPRRR) | | $L = \begin{bmatrix} 8 & 2 & 0 & 0 & 0 & 0 \\ 2 & 8 & 0 & 0 & 0 & 0 \\ 0 & 0 & 9 & 0 & 0 & 0 \\ 0 & 0 & 0 & 8 & 2 & 2 \\ 0 & 0 & 0 & 2 & 8 & 2 \\ 0 & 0 & 0 & 2 & 2 & 8 \end{bmatrix}$ | $M_{Ld} = \begin{bmatrix} 3 & 0 & 0 & 0 & 0 & 0 \\ 3 & 0 & 0 & 0 & 0 & 0 \end{bmatrix}$ |

*Definition of POC matrix of a leg*

The POC matrix of a leg should include not only the dimension of output, but also the direction of output. So, a kind of 2×f matrix (M) is proposed to map the POC of the leg to its topology matrix, that is

$$M = \begin{bmatrix} t_1 & L & t_f \\ r_1 & L & r_f \end{bmatrix} \quad (9)$$

The rules of the POC matrix of leg are as follows

① $f$ is the number of the single-DOF joints in a leg.

② $t_i$ or $r_i$ (i=1~f) denotes translation and rotation output respectively, and values is 0, 1, 2 or 3.

③ $t_i$ or $r_i \neq 0$ indicates the existence of independent motion.

④ when $t_i$ or $r_i$=3, the dimension of translation/rotation output is 3, and the direction is arbitrary in space and need not to be specified.

⑤ The dimension of translation/rotation is $\xi_r = \sum_{i=1}^{n} r_i$ / $\xi_t = \sum_{i=1}^{n} t_i$, and the total dimension of output at the end of the leg is $\xi = \xi_r + \xi_t$.

⑥ Output direction of column *i* is related to the axes of the $i^{th}$ joint in the leg.
   a. $t_i$=1 means there is an independent translation along the axis of the $i^{th}$ joint ($r_i$=0) or along the normal plane of the $i^{th}$ joint ($r_i$=1).
   b. when $t_i$ is 2, there are two independent translation in the normal plane of the $i^{th}$ joint axis.

*POC matrix of sub-chain*

Planar sub-chain (2-DOF $G_2$ and 3-DOF $G_3$) or spherical sub-chain (2-DOF $S_2$ and 3-DOF $S_3$) constitute a leg or part of a leg. Based on the definition of POC matrix, there are 7 types matrices for 15 sub-chains, as shown in Table 2.

TABLE 2 15 TYPES OF SUB-CHAINS AND THEIR POC

| No. | Type | Symbol | Dimension | Topology Matrix | POC matrix |
|---|---|---|---|---|---|
| 1 | | R∥R | 2 | $\begin{bmatrix} 8 & 1 \\ 1 & 8 \end{bmatrix}$ | $\begin{bmatrix} 1 & 0 \\ 1 & 0 \end{bmatrix}$ |
| 2 | $G_2$ | R⊥P | 2 | $\begin{bmatrix} 8 & 2 \\ 2 & 9 \end{bmatrix}$ | $\begin{bmatrix} 1 & 0 \\ 1 & 0 \end{bmatrix}$ |
| 3 | | P⊥R | 2 | $\begin{bmatrix} 9 & 2 \\ 2 & 8 \end{bmatrix}$ | $\begin{bmatrix} 0 & 1 \\ 0 & 1 \end{bmatrix}$ |
| 4 | | R//R//R | 3 | $\begin{bmatrix} 8 & 1 & 1 \\ 1 & 8 & 1 \\ 1 & 1 & 8 \end{bmatrix}$ | $\begin{bmatrix} 2 & 0 & 0 \\ 1 & 0 & 0 \end{bmatrix}$ |
| 5 | | R//R⊥P | 3 | $\begin{bmatrix} 8 & 1 & 2 \\ 1 & 8 & 2 \\ 2 & 2 & 9 \end{bmatrix}$ | $\begin{bmatrix} 2 & 0 & 0 \\ 1 & 0 & 0 \end{bmatrix}$ |
| 6 | | P⊥R//R | 3 | $\begin{bmatrix} 9 & 2 & 2 \\ 2 & 8 & 1 \\ 2 & 1 & 8 \end{bmatrix}$ | $\begin{bmatrix} 0 & 2 & 0 \\ 0 & 1 & 0 \end{bmatrix}$ |
| 7 | $G_3$ | R(⊥P)//R | 3 | $\begin{bmatrix} 8 & 2 & 1 \\ 2 & 9 & 2 \\ 1 & 2 & 8 \end{bmatrix}$ | $\begin{bmatrix} 2 & 0 & 0 \\ 1 & 0 & 0 \end{bmatrix}$ |
| 8 | | R(⊥P)⊥P | 3 | $\begin{bmatrix} 8 & 2 & 2 \\ 2 & 9 & 2 \\ 2 & 2 & 9 \end{bmatrix}$ | $\begin{bmatrix} 2 & 0 & 0 \\ 1 & 0 & 0 \end{bmatrix}$ |
| 9 | | P(⊥P)⊥R | 3 | $\begin{bmatrix} 9 & 2 & 2 \\ 2 & 9 & 2 \\ 2 & 2 & 8 \end{bmatrix}$ | $\begin{bmatrix} 0 & 0 & 2 \\ 0 & 0 & 1 \end{bmatrix}$ |
| 10 | | P(⊥R)⊥P | 3 | $\begin{bmatrix} 9 & 2 & 2 \\ 2 & 8 & 2 \\ 2 & 2 & 9 \end{bmatrix}$ | $\begin{bmatrix} 0 & 2 & 0 \\ 0 & 1 & 0 \end{bmatrix}$ |
| 11 | | R-R | 2 | $\begin{bmatrix} 8 & 0 \\ 0 & 8 \end{bmatrix}$ | $\begin{bmatrix} 0 & 0 \\ 1 & 1 \end{bmatrix}$ |
| 12 | $S_2$ | R⊥R(U) | 2 | $\begin{bmatrix} 8 & 2 \\ 2 & 8 \end{bmatrix}$ | $\begin{bmatrix} 0 & 0 \\ 1 & 1 \end{bmatrix}$ |
| 13 | $S_3$ | $R_1$-$R_2$-$R_3$ | 3 | $\begin{bmatrix} 8 & 0 & 0 \\ 0 & 8 & 0 \\ 0 & 0 & 8 \end{bmatrix}$ | $\begin{bmatrix} 0 & 0 & 0 \\ 1 & 1 & 1 \end{bmatrix}$ |

| No. | Type | Symbol | Dimension | Topology Matrix | POC matrix |
|---|---|---|---|---|---|
| 14 | | $R_1, R_2, R_3(S)$ | 3 | $\begin{bmatrix} 8 & 5 & 5 \\ 5 & 8 & 5 \\ 5 & 5 & 8 \end{bmatrix}$ | $\begin{bmatrix} 0 & 0 & 0 \\ 1 & 1 & 1 \end{bmatrix}$ |
| 15 | | $R_1 \perp R_2 \perp R_3$ ; $R_1 \perp R_3$ | 3 | $\begin{bmatrix} 8 & 2 & 2 \\ 2 & 8 & 2 \\ 2 & 2 & 8 \end{bmatrix}$ | $\begin{bmatrix} 0 & 0 & 0 \\ 1 & 1 & 1 \end{bmatrix}$ |

*POC Matrix of legs*

Take the leg UP as an example, it contains $R_1 \perp R_2 \perp P_3$. Dimension of the output is 3 (2-rotation and 1-translation), and the rotation is around $R_1$ axis and $R_2$ axis, and the translation along $P_3$ axis. Thus, POC matrix of this leg is $M_{La} = \begin{bmatrix} 0 & 0 & 1 \\ 1 & 1 & 0 \end{bmatrix}$.

**Description for PMs and its POC**

*Description for PMs*

To make information concise, a new structural representation of PMs is proposed, which consists of legs and joints on the two platforms.

Firstly, legs of a PM are labeled in turn with the numbers 1~6. Then, all legs are represented using the ordered topology matrix L. Finally, the joints on two platforms are regarded as two correspond virtual legs, expressed by the ordered topology matrices M and B. Therefore, the PM can be described as

PM=(L1, L2, ..., L(v+1), M, B)    (10)

The PM in Fig.2(A) has 3-UPS legs and 1-UP leg. The topological structure is PM=($L_1, L_2, L_3, L_4, M, B$). The joints on the two platforms are arbitrary.

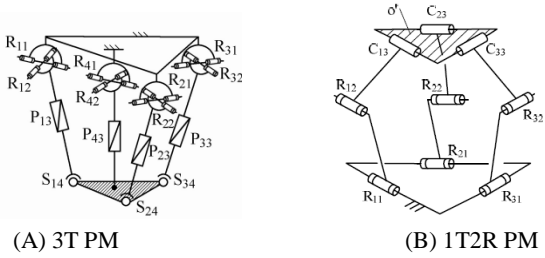

(A) 3T PM           (B) 1T2R PM
FIGURE 2 TWO KINDS OF PMS

The topology matrix $L_i$ (i=1~3) for 3-UPS legs and matrix $L_4$ for UP leg are respectively

$$L_i = \begin{bmatrix} 8 & 2 & 0 & 0 & 0 & 0 \\ 2 & 8 & 0 & 0 & 0 & 0 \\ 0 & 0 & 9 & 0 & 0 & 0 \\ 0 & 0 & 0 & 8 & 2 & 2 \\ 0 & 0 & 0 & 2 & 8 & 2 \\ 0 & 0 & 0 & 2 & 2 & 8 \end{bmatrix}, \quad L_4 = \begin{bmatrix} 8 & 2 & 2 \\ 2 & 8 & 2 \\ 2 & 2 & 9 \end{bmatrix}$$

Matrix M and B are

$$M = \begin{bmatrix} 8 & 0 & 0 & 0 \\ 0 & 8 & 0 & 0 \\ 0 & 0 & 8 & 0 \\ 0 & 0 & 0 & 8 \end{bmatrix}, \quad B = \begin{bmatrix} 8 & 0 & 0 & 0 \\ 0 & 8 & 0 & 0 \\ 0 & 0 & 8 & 0 \\ 0 & 0 & 0 & 8 \end{bmatrix}$$

*POC matrix of PMs*

According to the intersection property of sets, the direction of translation/rotation output of a PM is related to one or more joints in a certain leg. Thus the No. of this leg should be recorded. So, a column matrix is defined as

$$M_{pa} = \begin{bmatrix} t_1 & L & t_n \\ r_1 & L & r_n \end{bmatrix} \begin{bmatrix} l_1 \\ l_2 \end{bmatrix}, (n \leq 6) \quad (11)$$

where, $l_1$ and $l_2$ are the No. of the leg specifying the direction of translation/rotating output, n is the dimension of independent output.

The PM shown in Fig.2(A) has 2-rotation and 1-translation. The POC matrix is $M_{pa} = \begin{bmatrix} 0 & 0 & 1 \\ 1 & 1 & 0 \end{bmatrix} \begin{bmatrix} 4 \\ 4 \end{bmatrix}$, the double 4 indicate that the PM has a translation along $P_{43}$ joints, and the directions of rotation are around $R_{41}$ and $R_{42}$, both in the fourth leg.

## AUTOMATIC GENERATION OF POC AND DIMENSION OF INDEPENDENT DISPLACEMENT EQUATIONS

**Automatic POC generation of legs**

According to Eq.(3), POC of a leg is the union operation of the POC of all joints. Each leg can be considered as a serial of independent sub-chains and single-DOF joints. Thus, the sub-chains are extracted, and union operation of POCs are mapped to logical "OR" operation of matrices.

For convenience of programming, the POC matrix with dimension less than 6 is supplemented to the same dimension of 2×6

$$M_L = \begin{bmatrix} t_1 & t_2 & t_3 & t_4 & t_5 & t_6 \\ r_1 & r_2 & r_3 & r_4 & r_5 & r_6 \end{bmatrix} \quad (12)$$

Thus, the POC of a leg can be computed by

$$POC(M_L) = \sum POC(M_{G_3} + M_{G_2} + M_{S_3} + M_{S_2} + M_J) \quad (13)$$

where, $M_{G3}, M_{G2}, M_{S3}, M_{S2}$ and $M_J$ are the supplemented forms of the POC matrices of $G_3, G_2, S_3, S_2$ and single-DOF joints.

Here, based on the topology matrices of legs and the extracted sub-chains and single-DOF joints, the POC of the leg can be generated as follows

① Recognize topology matrix of the leg, and extract $G_3, G_2, S_3, S_2$ successively.
② Write out the supplemented matrix of $G_3, G_2, S_3, S_2$ and single-DOF joints.
③ Carry out "OR" operation on supplemented matrices of $G_3, G_2, S_3, S_2$ and single-DOF joints, and obtain the initial POC matrix of the leg, which denoted as $M_L^0 = \begin{bmatrix} t_1 & t_2 & t_3 & t_4 & t_5 & t_6 \\ r_1 & r_2 & r_3 & r_4 & r_5 & r_6 \end{bmatrix}$, $\sum_{i=1}^{6} r_i = \xi_r^0$, $\sum_{i=1}^{6} t_i = \xi_t^0$.
④ Process the translational output generated by parallel revolute joints.
  a. If $\xi_r^0 > 3$, then $\xi_r = 3$ and $\xi_t = \xi_t^0 + \xi_r^0 - 3$, and the translation newly generated lies in the normal plane of the single-DOF or $S_2$ rotational joints.
  b. If $\xi_r^0 \leq 3$ and $\xi_t^0 \leq 3$, the final POC matrix is $M_L = M_L^0$.

Table 1. gives the matrices of four typical legs. Here take the leg P⊥R//R⊥R//R as an example.

① Recognize and sequentially extract planar and spherical sub-chains: one $G_3$ sub-chain P⊥R//R and one $G_2$ planar sub-chaines R//R.
② Generate the supplemented POC matrix of $G_3$ and $G_2$ as $M_{G3} = \begin{bmatrix} 0 & 2 & 0 & 0 & 0 & 0 \\ 0 & 1 & 0 & 0 & 0 & 0 \end{bmatrix}$, $M_{G2} = \begin{bmatrix} 0 & 0 & 0 & 1 & 0 & 0 \\ 0 & 0 & 0 & 1 & 0 & 0 \end{bmatrix}$.
③ Carry out "or" operation of $M_{G3}$ and $M_{G2}$, and obtain the POC: $M_L = \begin{bmatrix} 0 & 2 & 0 & 1 & 0 & 0 \\ 0 & 1 & 0 & 1 & 0 & 0 \end{bmatrix} = \begin{bmatrix} 3 & 0 & 0 & 0 & 0 & 0 \\ 0 & 1 & 0 & 1 & 0 & 0 \end{bmatrix}$, $\xi_r^0=2$, and $\xi_t^0=3$.
④ $\xi_t^0 = 3$ and $\xi_r^0 = 2$, so $M_L = M_L^0$, that means $M_L = \begin{bmatrix} 3 & 0 & 0 & 0 & 0 & 0 \\ 0 & 1 & 0 & 1 & 0 & 0 \end{bmatrix}$, and $\xi_t = 3$ indicates that the leg has three independent translation outputs, $\xi_r=2$ indicates that it has two rotation outputs, $r_2=r_4=1$ means the two rotations are around the axes of $R_2$ and $R_4$ respectively.

**Automatic POC generation of PMs**

According to Eq.(4), the POC of a PM is the intersection operation on all legs in the PM. Among them, the translational output of the PM is the intersection of translational outputs of all legs, and the same goes for the rotating outputs.

Given two legs $L_1$ and $L_2$, numbered as $l_1$ and $l_2$, the POC matrices are $M_{Li} = \begin{bmatrix} t_{i1} & t_{i2} & t_{i3} & t_{i4} & t_{i5} & t_{i6} \\ r_{i1} & r_{i2} & r_{i3} & r_{i4} & r_{i5} & r_{i6} \end{bmatrix}$ ($i=1,2$), the dimension of translation/rotating outputs are $\xi_{ri} = \sum_{j=1}^{6} r_{ij}$ and $\xi_{ti} = \sum_{j=1}^{6} t_{ij}$.

For convenience of description, the translational output of leg i is denoted as $G_i=(\xi_{ti}, e_i)$ and the rotating output as $H_i=(\xi_{ri}, s_i)$, where $e_i/s_i$ are the direction of translational/rotating output of this leg. If $\xi_{ti}=0/\xi_{ri}=0$, $e_i/s_i$ are $\Phi$(empty set); if $\xi_{ti}=1/\xi_{ri}=1$, $e_i/s_i$ are spatial lines; if $\xi_{ti}=2/\xi_{ri}=2$, $e_i/s_i$ are plane; if $\xi_{ti}=3/\xi_{ri}=3$, $e_i/s_i$ are "-".

Suppose the intersection result is denoted as $M_{L1 \cap L2} = \begin{bmatrix} t_1 & t_2 & t_3 & t_4 & t_5 & t_6 \\ r_1 & r_2 & r_3 & r_4 & r_5 & r_6 \end{bmatrix}$, the dimension of translational/rotating output is $\sum_{i=1}^{6} r_i / \sum_{i=1}^{6} t_i$ respectively. The translational output is $G_{(1\cap2)}=(\xi_t, e)$ and the rotating output is $H_{(1\cap2)}=(\xi_r, s)$, where e and s are the direction correspond.

Note that the legs $L_1$ and $L_2$, which participate in computing, can be either the legs constituting the PM or the sub-PM composed by several legs. The POC of PMs can be calculated as follows.

*Algorithm for translational output of PMs*

The intersection operation of translational output of two legs is to solve $G_{(1\cap2)}=G_1 \cap G_2$, which can be obtained according the rules shown in Table 3.

TABLE 3 INTERSECTION RULES FOR TRANSLATION

| $L_2 \backslash L_1$ | $\xi_{t1}=0$ | $\xi_{t1}=1$ | $\xi_{t1}=2$ | $\xi_{t1}=3$ |
|---|---|---|---|---|
| $\xi_{t2}=0$ | $G_2$ | $G_2$ | $G_2$ | $G_2$ |
| $\xi_{t2}=1$ | $G_1$ | $e_1\|\|e_2$  $G_2$ <br> $e_1\not\|\|e_2$ $\xi_t=0,e=\Phi$ | $e_1\|\|e_2$  $G_1$ <br> $e_1\not\|\|e_2$ $\xi_t=0,e=\Phi$ | $G_2$ |
| $\xi_{t2}=2$ | $G_1$ | $e_1\|\|e_2$  $G_1$ <br> $e_1\not\|\|e_2$ $\xi_t=0, e=\Phi$ | $e_1\|\|e_2$  $G_2$ <br> $e_1\not\|\|e_2$ $\xi_t=1,e=e_1\cap e_2$ | $G_2$ |
| $\xi_{t2}=3$ | $G_1$ | $G_1$ | $G_1$ | $G_1$ |

*Algorithms for rotating output of PMs*

Similarly, the intersection of rotating output of two POC matrices is to solve $H_{(1\cap2)}=H_1 \cap H_2$. The intersection rules are listed in Table 4.

TABLE 4 INTERSECTION RULES FOR ROTATION

| $L_2 \backslash L_1$ | $\xi_{r1}=0$ | $\xi_{r1}=1$ | $\xi_{r1}=2$ | $\xi_{r1}=3$ |
|---|---|---|---|---|
| $\xi_{r2}=0$ | $H_2$ | $H_2$ | $H_2$ | $H_2$ |
| $\xi_{r2}=1$ | $H_1$ | $s_1\|\|s_2$  $H_2$ <br> $s_1\not\|\|s_2$ $\xi_r=0,s=\Phi$ | $s_1\|\|s_2$  $H_1$ <br> $s_1\not\|\|s_2$ $\xi_r=0,s=\Phi$ | $H_2$ |
| $\xi_{r2}=2$ | $H_1$ | $s_1\|\|s_2$  $H_1$ <br> $s_1\not\|\|s_2$ $\xi_r=0,s=\Phi$ | $s_1\|\|s_2$  $H_2$ <br> $s_1\not\|\|s_2$ $\xi_t=1,s=s_1\cap s_2$ | $H_2$ |
| $\xi_{r2}=3$ | $H_1$ | $H_1$ | $H_1$ | $H_1$ |

Accordance to the analysis above, the flow of generating POC matrices of PMs are as follows
① Input the number of legs, and POC of each leg.
② Intersection operation on translational outputs.
③ Intersection operation on rotating outputs.
④ Output the result of POC of the PM.

Here take the PM of 3-RRC as an example. As shown in Fig.2(B). The topology matrices of legs, joints on the two platforms are described previously.
① Input the number of legs, and POC of each leg.
Number of legs is 3, and topology matrices of legs are

$M_{Li} = \begin{bmatrix} 2 & 0 & 0 & 0 & 0 & 0 \\ 1 & 0 & 0 & 0 & 0 & 0 \end{bmatrix} + \begin{bmatrix} 0 & 0 & 0 & 1 & 0 & 0 \\ 0 & 0 & 0 & 0 & 0 & 0 \end{bmatrix} = \begin{bmatrix} 3 & 0 & 0 & 0 & 0 & 0 \\ 1 & 0 & 0 & 0 & 0 & 0 \end{bmatrix} = \begin{bmatrix} G_i \\ H_i \end{bmatrix}$ (i=1,2,3).

Thus, $\xi_{ti}=3$ means arbitrary translation in space, and $\xi_{ri}=1$ is one-dimension rotation around the $R_{i1}$ axis in $i^{th}$ leg, i.e. $s_i=R_{i1}$.
② Intersection operation on translational outputs
According to Table 3, $\xi_{ti}=3$, then $\xi_{L1t}=3$, and $G_{(1\cap2)}=G_2$.
③ Intersection operation on rotating outputs
According to Table 4, $\xi_{r1}=\xi_{r2}=\xi_{r3}=1$, $s_1=R_{11}$, $s_2=R_{21}$, $s_3=R_{31}$ and $s_1\not\|s_2\not\|s_3$, then $\xi_r=0$, $s=\Phi$.

④ Output the result of POC of this PM

$$M_{pa}=M_{pa(1\cap2)}\cap M_{b3}=\begin{bmatrix}G_2\\\phi\end{bmatrix}\cap\begin{bmatrix}G_3\\H_3\end{bmatrix}=\begin{bmatrix}G_3\\\phi\end{bmatrix}$$

The output of 3-RRC PM is 3-translation, and the result is consistent with those in Ref [5].

**Calculating the number of independent displacement equations**

Eq(6) shows that solving the number of independent displacement equations of an independent loop (SLC) involves the union operation of POC matrices. For the legs $L_1$ and $L_2$ mentioned above, supposing that the result matrix of union is

$$M_{L1\cup L2}=\begin{bmatrix}t_1 & t_2 & t_3 & t_4 & t_5 & t_6\\r_1 & r_2 & r_3 & r_4 & r_5 & r_6\end{bmatrix}$$

The dimension of independent output is

$$\xi_{Lr}=\sum_{i=1}^{6}r_i \qquad \xi_{Lt}=\sum_{i=1}^{6}t_i$$

Translational output is $G_{(1\cup2)}=(\xi_{Lt}, e)$ and rotating output is $H_{(1\cup2)}=(\xi_{Lr}, s)$, where e/s is the direction of translation/rotation. Then the number of the independent displacement equations of the independent loop is $\xi_L=\xi_{Lt}+\xi_{Lr}$.

*Algorithms for the dimension of translation output of SLC*

Solving the dimension of translational output of a SLC is essentially to calculate $\xi_{Lt}=\dim(G_1\cup G_2)$, whose operation rules are shown in Table 5.

TABLE 5. ALGORITHMS FOR THE DIMENSION OF TRANSLATIONAL OUTPUT

| L₁ / L₂ | $\xi_{t1}=0$ | $\xi_{t1}=1$ | | $\xi_{t1}=2$ | | $\xi_{t1}=3$ |
|---|---|---|---|---|---|---|
| $\xi_{t2}=0$ | 0 | 1 | | 2 | | 3 |
| $\xi_{t2}=1$ | 1 | $e_1\|e_2$ | $\xi_{Lt}=1$ | $e_1\|e_2$ | $\xi_{Lt}=2$ | 3 |
| | | $e_1\nparallel e_2$ | $\xi_{Lt}=2$ | $e_1\nparallel e_2$ | $\xi_{Lt}=3$ | |
| $\xi_{t2}=2$ | 2 | $e_1\|e_2$ | $\xi_{Lt}=2$ | $e_1\|e_2$ | $\xi_{Lt}=2$ | 3 |
| | | $e_1\nparallel e_2$ | $\xi_{Lt}=3$ | $e_1\nparallel e_2$ | $\xi_{Lt}=3$ | |
| $\xi_{t2}=3$ | 3 | 3 | | 3 | | 3 |

Similarly, the dimension of rotating output of a SLC is to calculate $\xi_{Lr}=\dim(H_1\cup H_2)$, and its operation rules are shown in Table 6.

TABLE 6. ALGORITHMS FOR THE DIMENSION OF ROTATING OUTPUT

| L₁ / L₂ | $\xi_{t1}=0$ | $\xi_{t1}=1$ | | $\xi_{t1}=2$ | | $\xi_{t1}=3$ |
|---|---|---|---|---|---|---|
| $\xi_{t2}=0$ | 0 | 1 | | 2 | | 3 |
| $\xi_{t2}=1$ | 1 | $s_1\|s_2$ | $\xi_{Lr}=1$ | $s_1\|s_2$ | $\xi_{Lr}=2$ | 3 |
| | | $s_1\nparallel s_2$ | $\xi_{Lr}=2$ | $s_1\nparallel s_2$ | $\xi_{Lr}=$ | |
| $\xi_{t2}=2$ | 2 | $s_1\|s_2$ | $\xi_{Lr}=2$ | $s_1\|s_2$ | $\xi_{Lr}=2$ | 3 |
| | | $s_1\nparallel s_2$ | $\xi_{Lr}=3$ | $s_1\nparallel s_2$ | $\xi_{Lr}=3$ | |
| $\xi_{t2}=3$ | 3 | 3 | | 3 | | 3 |

**PROCEDURE OF AUTOMATIC MOBILITY ANALYSIS OF PMS**

Based on the algorithm for POC of legs and PMs and for the number of the independent displacement equations, the mobility analysis flow for PMs is established as shown in Fig.3.

**Step1.** Input topological structure of the PM.
Input topology matrices $L_1, \ldots, L_{(v+1)}$ and number leg in sequence automatically, and input the axis relation matrix M and B on the moving platform and fixed platform. Obtain the number of the joints $f_i$ and the number of $v+1$ legs.

**Step2.** Calculate $(v+1)$ POC matrix of legs: $M_{L1},\ldots M_{L(v+1)}$.

**Step3.** Calculate the sum of all joints in the PM: $f_{(1\sim(v+1))}=\sum_{i=1}^{v+1}f_i$.

**Step4.** Calculate the number of the independent displacement equations of the 1st loop $SLC_1$: $\xi_1$
① Calculate the dimension of the independent translational output of $SLC_1$: $\xi_{L1t}=\dim(G_1\cup G_2)$.
② Calculate the dimension of te independent rotating output of $SLC_1$: $\xi_{L1r}=\dim(H_1\cup H_2)$.
③ Calculate $\xi_{L1}=\xi_{L1t}+\xi_{L1r}$.

**Step5.** Calculate the POC matrix $MP_{(1\cap2)}$ of the sub-PM $P_{(1-2)}$ composed by the 1th and 2th legs.
① Calculate the translational output $G_{(1\cap2)}=G_1\cap G_2$.
② Calculate the rotating output $H_{(1\cap2)}=H_1\cap H_2$.
③ Get $MP_{(1\cap2)}=\begin{bmatrix}G_{(1\cap2)}\\H_{(1\cap2)}\end{bmatrix}$.

**Step6.** Calculate the POC matrix $M_{pa(1\sim j)}=M_{pa(1\sim(j-1))}\cap M_{bj}$ of the sub-PM $P_{(1-j)}$ composed by the front $j^{th}$ (j=3,...,v) legs.

**Step7.** Calculate the number of the independent displacement equations of the $j^{th}$ (j=2,...,v) independent loop $SLC_j$: $\xi_j=\dim(M_{pa(1\sim j)}\cup M_{b(j+1)})$.

**Step8.** Calculate the number of DOF: $F=f_{(1\sim(v+1))}-\sum_{i=1}^{v}\xi_i$.

**Step9.** Determine the property of mobility of this PM: $M_{pa}=M_{pa(1-v)}\cap M_{L(v+1)}$.

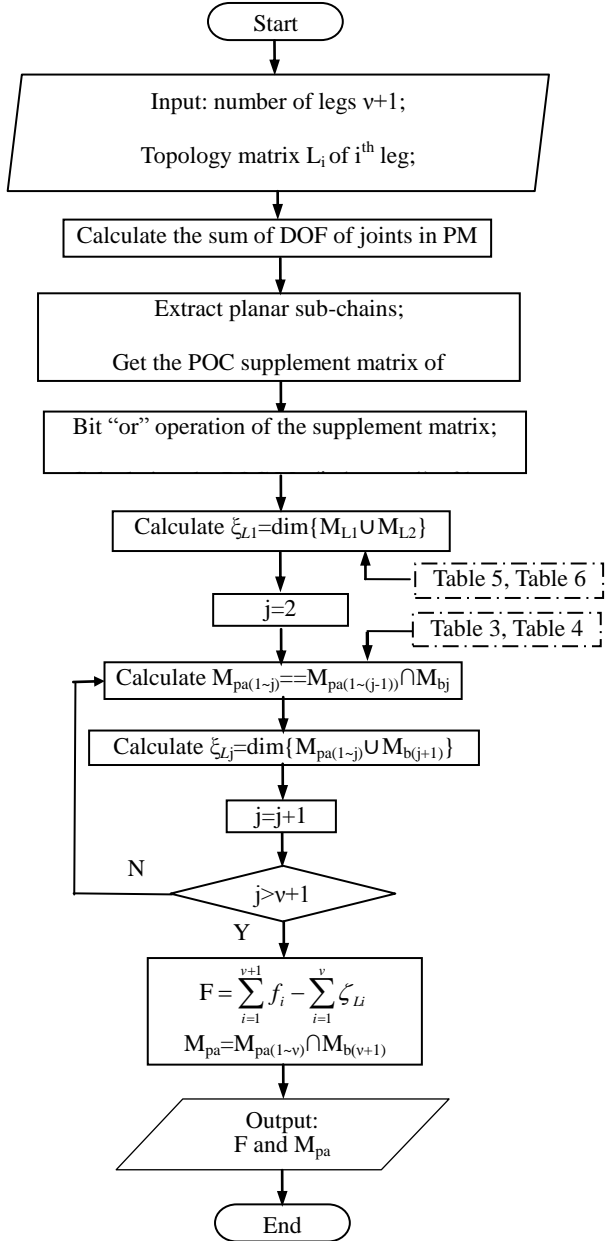

FIGURE 3 THE FLOW CHART OF MOBILITY ANALYSIS BASED ON POC

## SOFTWARE IMPLEMENTATION AND CASE STUDY

### Software implementation

According to the basin principle discussed above, an automatic mobility analysis platform for PMs was developed by VC++ 6.0. Using the platform, large number of PMs listed in [6] have been analyzed automatically based on the input of topology matrix and results have proved the validity of the software.

The input information mainly includes the number of legs and topology matrix. Using decimal array and easy to operate.

The analysis results include the DOF, the dimension of the independent translational/rotating output of the moving platform, and their corresponding axis direction.

### Case study

Detailed examples of mobility analysis of two PMs are presented in this section.

*Mobility analysis of the Tricept PM*

The PM of Tricept shown in Fig.2(A) has 3 DOF which is one translation and two rotation. The legs are 3-UPS and 1-UP. The axes of joints on the two platforms are of general.

(1) Input topological structural matrices of the PM

Topology matrix of UPS leg is $L_i = \begin{bmatrix} 8 & 2 & 1 & 0 & 0 \\ 2 & 9 & 2 & 0 & 0 \\ 1 & 2 & 8 & 0 & 0 \\ 0 & 0 & 0 & 8 & 5 \\ 0 & 0 & 0 & 5 & 8 \end{bmatrix}$, (i=1~3).

Topology matrix of UP leg is $L_4 = \begin{bmatrix} 8 & 2 & 2 \\ 2 & 8 & 2 \\ 2 & 2 & 9 \end{bmatrix}$.

To the two virtual legs on the platforms, topology matrices M and B are $M = \begin{bmatrix} 8 & 0 & 0 & 0 \\ 0 & 8 & 0 & 0 \\ 0 & 0 & 8 & 0 \\ 0 & 0 & 0 & 8 \end{bmatrix}$, $B = \begin{bmatrix} 8 & 0 & 0 & 0 \\ 0 & 8 & 0 & 0 \\ 0 & 0 & 8 & 0 \\ 0 & 0 & 0 & 8 \end{bmatrix}$.

Number of legs $v+1=4$

(2) Calculate POC matrix of legs $M_{L1},...M_{L4}$.

① $M_{Li} = \begin{bmatrix} 3 & 0 & 0 & 0 & 0 \\ 3 & 0 & 0 & 0 & 0 \end{bmatrix} = \begin{bmatrix} G_i \\ H_i \end{bmatrix}$, $\xi_{ir}=3$, and $\xi_{it}=3$ (i=1~3)

② $M_{L4} = \begin{bmatrix} 0 & 0 & 1 & 0 & 0 & 0 \\ 1 & 1 & 0 & 0 & 0 & 0 \end{bmatrix} = \begin{bmatrix} G_4 \\ H_4 \end{bmatrix}$, and $\xi_{4t}=1$, $\xi_{4r}=2$.

(3) Get the sum of all joints in this PM: $f_{(1-4)}=6+6+6+3=21$.

(4) Calculate the number of the independent displacement equations of the first loop (SLC$_1$)

① $\xi_{ti}=3$, as shown in Table 5, $\xi_{L1t}=3$, $G_{(1\cup2)}=G_2$ (i=1,2).

② $\xi_{ri}=3$, then $\xi_{L1r}=3$, $H_{(1\cup2)}=H_2$ (i=1,2).

③ So, $\xi_{L1}=\xi_{L1t}+\xi_{L1r}=3+3=6$.

(5) Calculate the POC matrix $MP_{(1\cap2)}$ of sub-PM constituted by 1$^{st}$ and 2$^{nd}$ legs.

① $\xi_{ti}=3$, as shown in Table 3, $G_{(1\cap2)}=G_2$ (i=1,2).

② $\xi_{ri}=3$, as shown in Table 4, $H_{(1\cap2)}=H_2$ (i=1,2).

③ So, $MP_{(1\cap2)} = \begin{bmatrix} G_2 \\ H_2 \end{bmatrix}$.

(6) Calculate the POC matrix of sub-PM composed by the front three legs: $M_{pa(1\sim3)} = M_{pa(1\sim2)} \cap M_{b3} = \begin{bmatrix} G_2 \\ H_2 \end{bmatrix} \cap \begin{bmatrix} G_3 \\ H_3 \end{bmatrix} = \begin{bmatrix} G_3 \\ H_3 \end{bmatrix}$.

(7) Calculate the number of the independent displacement equations of SLC$_2$: $\xi_{L2}=\dim(M_{pa(1\sim2)} \cup M_{b3})$.

① $\xi_{t(1\cap2)}=3$, and $\xi_{t3}=3$, then $\xi_{L2t}=3$.

② $\xi_{r(1\cap2)}=3$, and $\xi_{r3}=3$, then $\xi_{L2r}=3$.

③ then, $\xi_{L2}=\xi_{L2t}+\xi_{L2r}=3+3=6$.

Similarly, the number of the independent displacement equations of $SLC_3$ is $\xi_{L3}=\xi_{L3t}+\xi_{L3r}=3+3=6$.

(8) Get the number of DOF: $F=f_{(1\sim4)}-\sum_{i=1}^{3}\xi_i=21-(6+6+6)=3$.

(9) Get the property of mobility:

$M_{pa}=M_{pa(1-3)}\cap M_{b4}=\begin{bmatrix}G_3\\H_3\end{bmatrix}\cap\begin{bmatrix}G_4\\H_4\end{bmatrix}=\begin{bmatrix}G_4\\H_4\end{bmatrix}$.

The results show that this mechanism has 1T2R output, the rotating directions are around $R_{41}$ and $R_{41}$, and the translational direction along $P_{43}$ joint. The result generated by software is shown in Fig.4, which is consistent with those in Ref.[6].

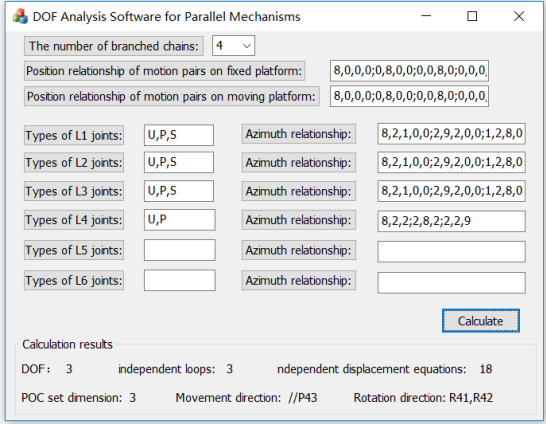

FIGURE 4. MOBILITY ANALYSIS OF TRICEPT

*Mobility analysis of a 3-RRC PM*

Fig.2(B) shows a 3-RRC PM. It has three identical RRC legs connecting the two platforms, labeled with the number 1~3. The topological structure of this PM is $PM=(L_1,L_2,L_3,M,B)$. The fixed platform is a triangle, and axes of the joints on the two platforms are all coplanar.

(1) Input topology matrices of the PM

Topology matrix of RRC leg is $L_i=\begin{bmatrix}8&1&1&1\\1&8&1&1\\1&1&8&1\\1&1&1&9\end{bmatrix}$, (i=1~3).

To two virtual legs on the platforms, topology matrices M and B are $M=\begin{bmatrix}9&5&5\\5&9&5\\5&5&9\end{bmatrix}$ and $B=\begin{bmatrix}8&5&5\\5&8&5\\5&5&8\end{bmatrix}$.

Number of legs $v+1=3$.

(2) Calculate POC matrix of legs $M_{L1},...M_{L3}$.

① Recognize and sequentially extract planar and spherical sub-chains: one $G_3$ sub-chain R//R//R and one single-DOF P joint.

② Generate the POC supplemented matrices of $G_3$ and P as $M_{G3}=\begin{bmatrix}2&0&0&0&0\\1&0&0&0&0\end{bmatrix}$ and $M_P=\begin{bmatrix}0&0&0&1&0&0\\0&0&0&0&0&0\end{bmatrix}$.

③ Carry out "OR" operation on POC matrices of all legs

$M_{Li}=\begin{bmatrix}2&0&0&1&0&0\\1&0&0&0&0&0\end{bmatrix}=\begin{bmatrix}3&0&0&0&0&0\\1&0&0&0&0&0\end{bmatrix}$, $\xi_r^0=1$, and $\xi_t^0=3$ (i=1~3)

④ $\xi_t=3$ and $\xi_r=1$, then $M_L=M_{La}^0$, that means

$M_{Li}=\begin{bmatrix}3&0&0&0&0&0\\1&0&0&0&0&0\end{bmatrix}$, and $\xi_{it}=3$ indicate that the leg has three translational output along arbitrary direction, $\xi_{ir}=1$ indicates that it has one rotating output, $r_{i1}=1$ means that the rotation is around the axes of $R_{i1}$, i.e. $s_i=R_{i1}$.

$M_{Li}=\begin{bmatrix}3&0&0&0&0&0\\1&0&0&0&0&0\end{bmatrix}=\begin{bmatrix}G_i\\H_i\end{bmatrix}$

(3) Calculate the sum of all joints in this PM: $f_{(1-3)}=4+4+4=12$.

(4) Calculate the number of the independent displacement equations of the first loop ($SLC_1$)
① $\xi_{ti}=3$, as shown in Table 5, $\xi_{L1t}=3$ (i=1,2).
② $\xi_{ri}=1$, and $s_1\not\parallel s_2$, then, $\xi_{L1r}=2$ (i=1,2).
③ So, $\xi_{L1}=\xi_{L1t}+\xi_{L1r}=3+2=5$.

(5) Calculate the POC matrix $MP_{(1\cap2)}$ of sub-PM constituted by the 1st and 2nd legs
① $\xi_{ti}=3$, as shown in Table 3, $G_{(1\cap2)}=G_2$ (i=1,2).
② $\xi_{ri}=1$, and $s_1\not\parallel s_2$, as shown in Table 4, $H_{(1\cap2)}=\Phi$ (i=1,2).
③ So, $MP_{(1\cap2)}=\begin{bmatrix}G_2\\\phi\end{bmatrix}$.

(6) Calculate the POC matrix of PM constituted by the front three legs: $M_{pa(1\sim3)}=M_{pa(1\sim2)}\cap M_{L3}=\begin{bmatrix}G_2\\\phi\end{bmatrix}\cap\begin{bmatrix}G_3\\H_3\end{bmatrix}=\begin{bmatrix}G_3\\\phi\end{bmatrix}$.

(7) Calculate the number of the independent displacement equations of $SLC_2$: $\xi_{L2}=\dim(M_{pa(1\sim2)}\cup M_{b3})$
① $\xi_{t3}=3$, as shown in Table 3, $\xi_{L2t}=3$.
② $\xi_{r(1-2)}=0$, and $\xi_{r3}=1$, as shown in Table 4, $\xi_{L2r}=1$.
③ Thus, $\xi_{L2}=\xi_{L2t}+\xi_{L2r}=3+1=4$.

(8) Get the number of DOF: $F=f_{(1\sim3)}-\sum_{i=1}^{3}\xi_i=12-(5+4)=3$

(9) Get the property of mobility: $M_{pa}=M_{pa(1-3)}=\begin{bmatrix}G_3\\\phi\end{bmatrix}$.

It shows that this PM has 3 DOF, three translation. The result automatically generated shown in Fig.5. The result is consistent with those in Ref[5].

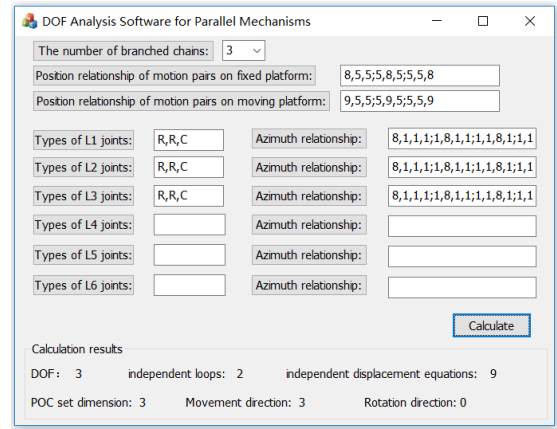

FIGURE 5. MOBILITY ANALYSIS OF 3-RRC PM

## CONCLUSIONS

A matrix description for mapping the topological structure with POC of legs and PMs was established. It includes not only the type and size of the translational/rotary output, but also represents the orientation of the output axis. By extracting the planar and spherical sub-chains orderly, the POC of a leg can be transformed into the logical "OR" operation of the matrices. An algorithm for POC of a PM was established without manual intervention. Algorithm of mobility analysis automatically of PMs is proposed. Software for mobility analysis of PMs was created and typical examples were provided in detail to show its effectiveness.

## ACKNOWLEDGMENTS:

This research is sponsored by the NSFC (Grant No. 51475050).